\documentclass[conference]{IEEEtran}
\IEEEoverridecommandlockouts
\usepackage{cite}
\usepackage{graphicx}
\usepackage{url}
\usepackage{amsmath}
\usepackage{amssymb}
\usepackage[caption = false, font=footnotesize]{subfig}
\usepackage{float}
\usepackage{booktabs} 
\usepackage{multirow}
\usepackage[colorlinks=true,linkcolor=blue]{hyperref}%
\usepackage{array}
\usepackage{color}
\usepackage{tabularx}
\usepackage{longtable}
\usepackage{tabu}
\usepackage{pifont}
\usepackage{amssymb}
\usepackage{ragged2e}
\usepackage{chemformula}

\begin{document}


\title{Rethinking Generative Human Video Coding with Implicit Motion Transformation
}

\author{%
Bolin Chen$^{\ast}$, Ru-Ling Liao$^{\ast}$, Jie Chen$^{\ast}$ and Yan Ye$^{\ast}$ \\[0.5em]
{\begin{minipage}{\linewidth}\begin{center}
\begin{tabular}{ccc}
\begin{tabular}{cccc}
\centering
$^{\ast}$Alibaba DAMO Academy \& Hupan Laboratory  & \hspace*{0.1in} 
\end{tabular}
 \\[0.5em]
E-mail: \{chenbolin.chenboli, ruling.lrl, jiechen.cj, yan.ye\}@alibaba-inc.com
\end{tabular}
\end{center}\end{minipage}}
}

\maketitle

\begin{abstract}
Beyond traditional hybrid-based video codec, generative video codec could achieve promising compression performance by evolving high-dimensional signals into compact feature representations for bitstream compactness at the encoder side and developing explicit motion fields as intermediate supervision for high-quality reconstruction at the decoder side. This paradigm has achieved significant success in face video compression. However, compared to facial videos, human body videos pose greater challenges due to their more complex and diverse motion patterns, \textit{i.e.,} when using explicit motion guidance for Generative Human Video Coding (GHVC), the reconstruction results could suffer severe distortions and inaccurate motion. As such, this paper highlights the limitations of explicit motion-based approaches for human body video compression and investigates the GHVC performance improvement with the aid of Implicit Motion Transformation, namely IMT. In particular, we propose to characterize complex human body signal into compact visual features and transform these features into implicit motion guidance for signal reconstruction. Experimental results demonstrate the effectiveness of the proposed IMT paradigm, which can facilitate GHVC to achieve high-efficiency compression and high-fidelity synthesis. 
\end{abstract}


\section{Introduction}

\begin{figure}[tb]
\centering
\subfloat[Existing Pipeline: Explicit Motion Estimation \& Warp-based Generation]{\includegraphics[width=0.45 \textwidth]{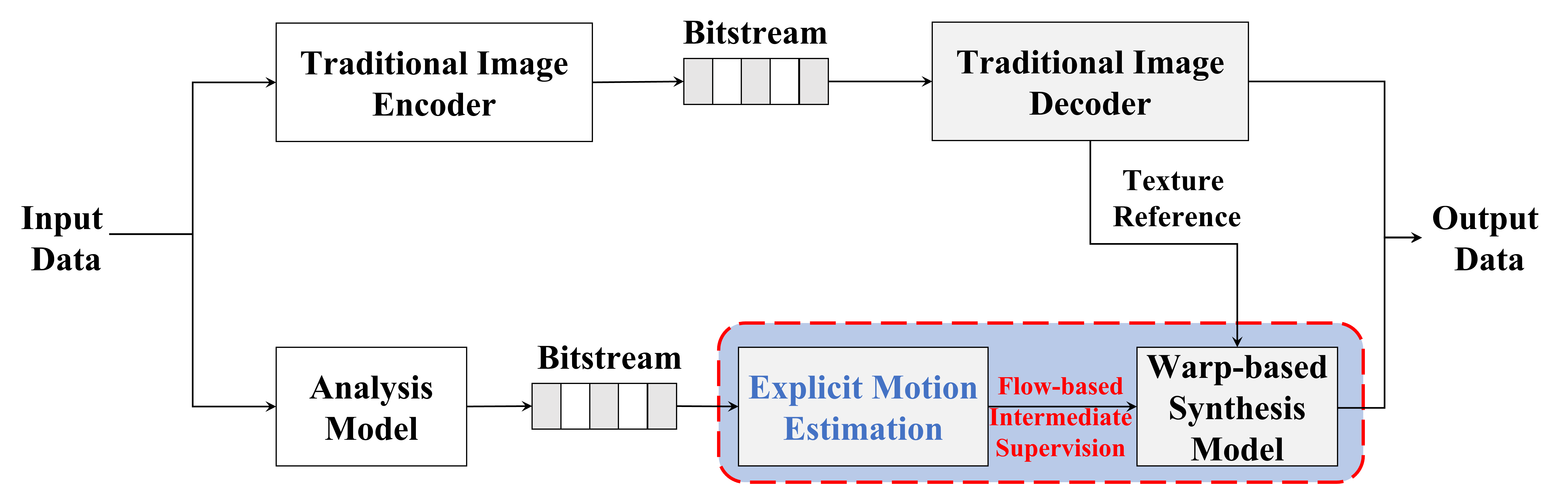}}
\hfil
\subfloat[Proposed Pipeline: Implicit Motion Transformation \& Direct Generation]{\includegraphics[width=0.45
\textwidth]{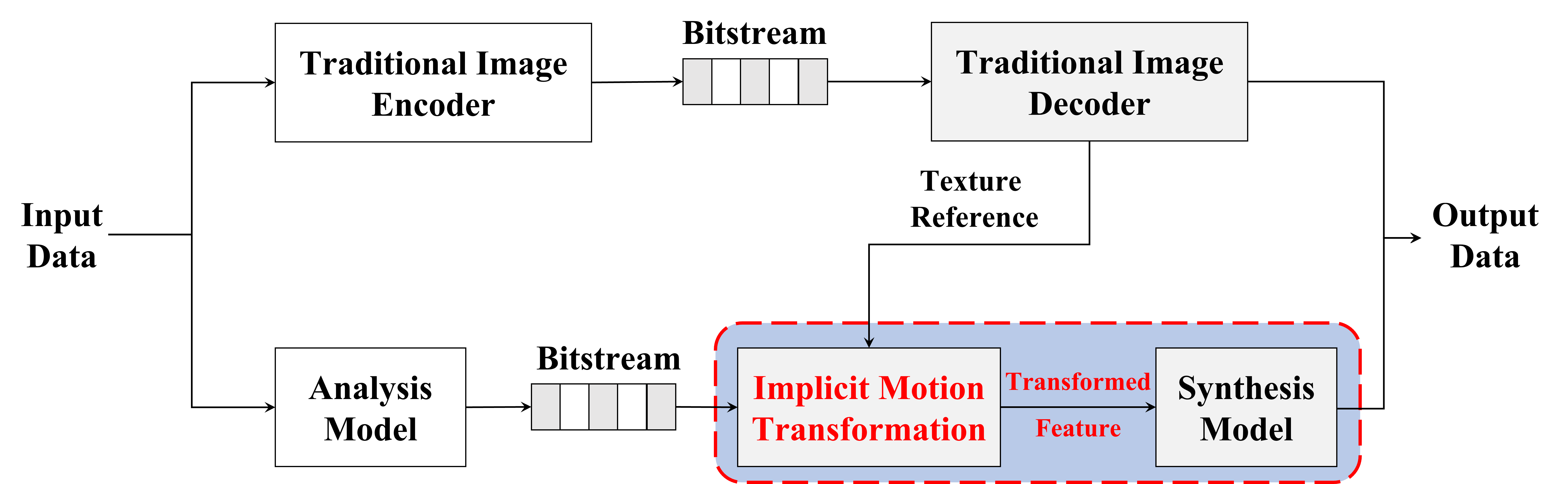}}
\vspace{-1em}
\subfloat[Visual Example Comparisons]{\includegraphics[width=0.4
\textwidth]{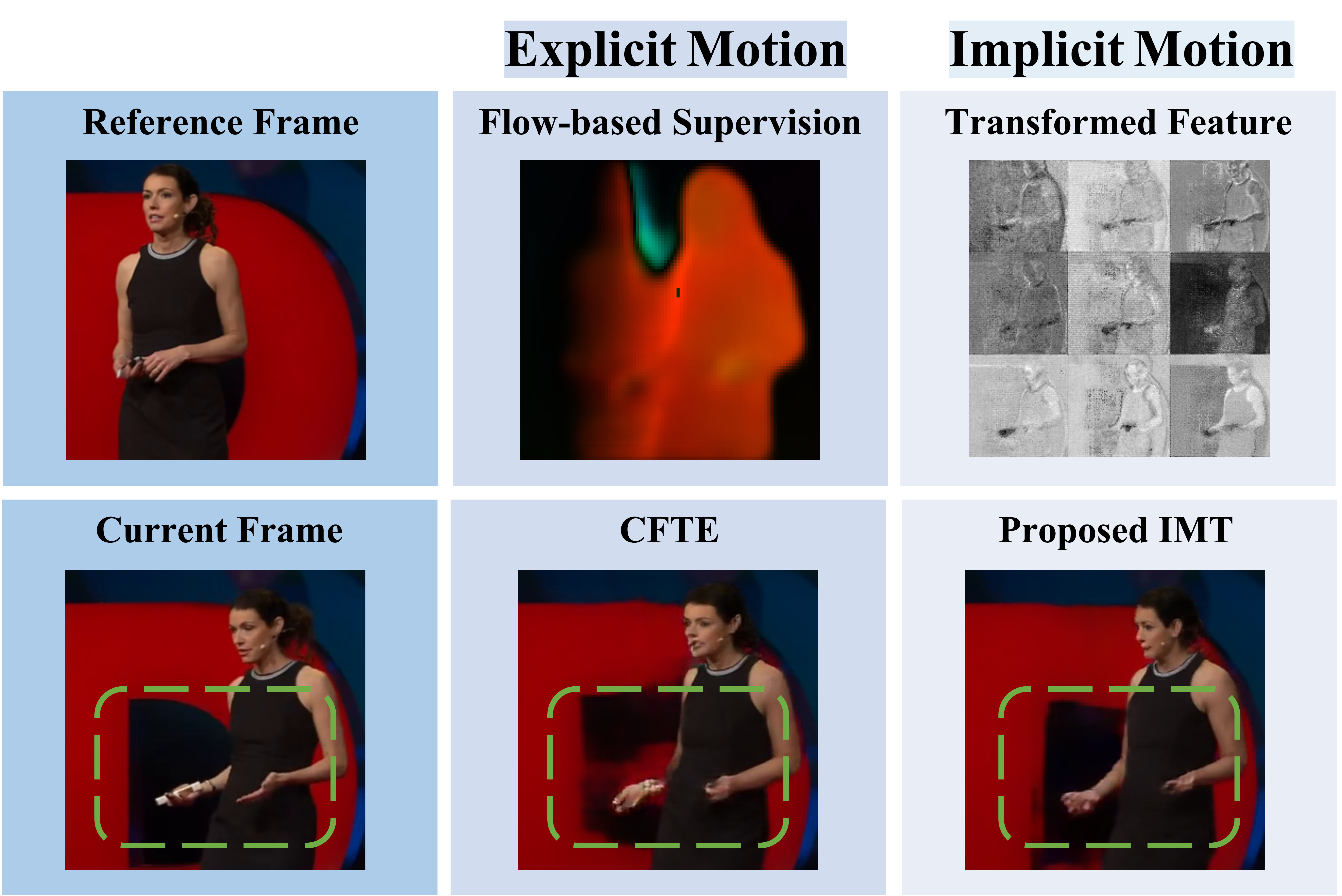}}
\caption{Comparisons of explicit motion estimation and implicit motion transformation paradigms for generative human video coding. Herein, CFTE~\cite{CHEN2022DCC} and the proposed IMT both utilize the same size of 6$\times$6 as transmitted symbols to describe human body motion, but leverage different motion estimation schemes for final signal reconstruction.  }
\vspace{-1.5em} 
\label{framework_comparison}  
\end{figure}

In recent years, generative video coding~\cite{10647820,10533752} has emerged as a powerful technique for compressing high-dimensional video signals at the feature level beyond traditional video codecs~\cite{wiegand2003overview,sullivan2012overview,bross2021overview}.
Generally, as illustrated in Fig.~\ref{framework_comparison} (a), these approaches could convert intricate video data into compact feature representations to enhance bitstream compactness at the encoder side, and rely on the strong inference capabilities of deep generative models~\cite{goodfellow2014generative,NEURIPS2021_49ad23d1} to evolve these compact representations into dense motion guidance for improved reconstruction quality at the decoder side. 
This paradigm has demonstrated remarkable success in compressing facial videos, leveraging their relatively compact but predictable feature patterns (\textit{e.g.,} 2D landmarks~\cite{9455985}, 2D keypoints~\cite{9414731}, 3D keypoints~\cite{FV2V,10811831}, compact feature~\cite{CHEN2022DCC,chen2023csvt}, progressive tokens~\cite{CHEN2025DCC} and facial semantics~\cite{11002361}).

\begin{figure*}[t]
\vspace{-2.2em}
\centering
\centerline{\includegraphics[width=0.97\textwidth]{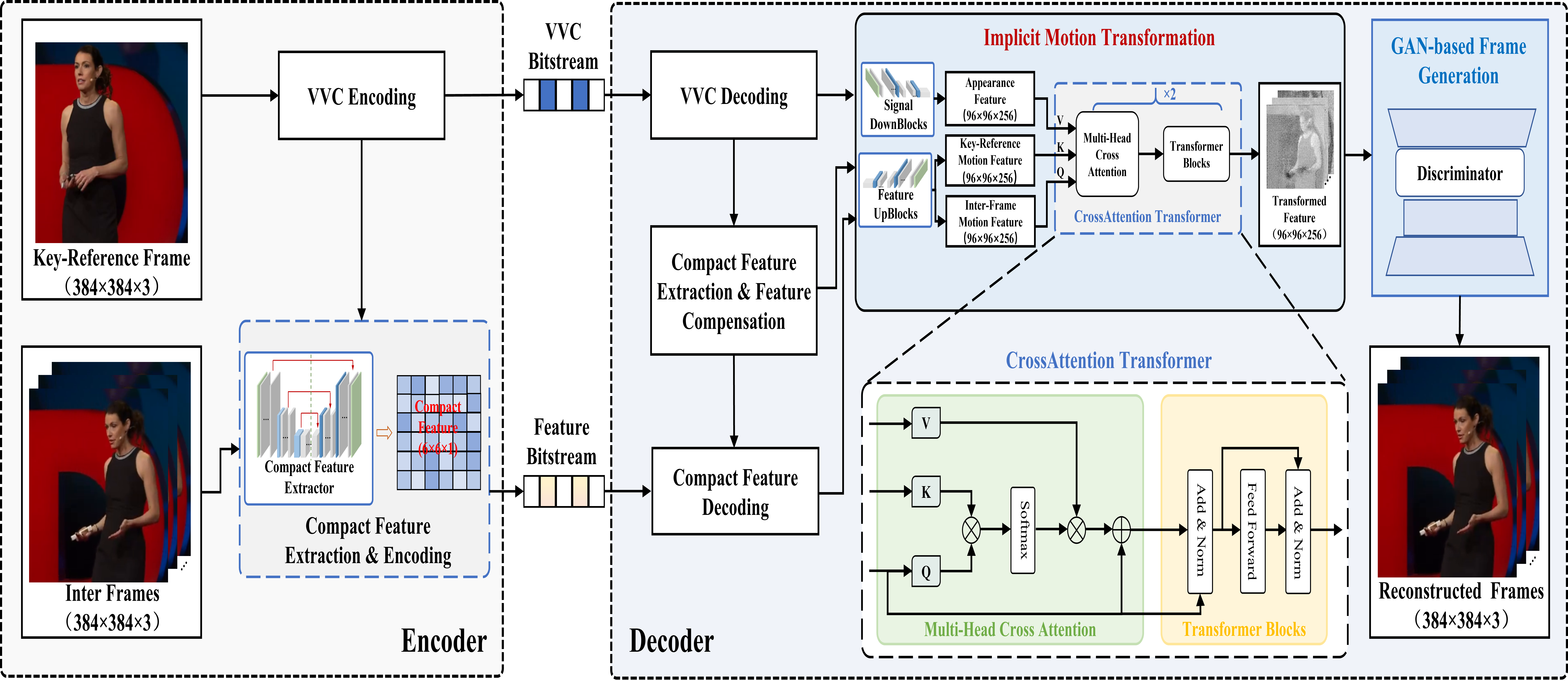}}
\caption{Overview of generative human video coding framework with the proposed IMT scheme towards high-efficiency human body video communication.}
\label{fig2}
\vspace{-1.5em}
\end{figure*}

However, such explicit motion estimation strategies face significant challenges when applied to generative video compression for the human body scenario, which exhibits complex and diverse motion dynamics. Human body movements are inherently more varied and intricate than facial expressions, leading to difficulties in achieving accurate reconstructions using explicit motion guidance and warp-based generation schemes. As described in~\cite{10656311}, the limitation of explicit motion guidance, such as optical flow, lies in its focus on capturing pixel-level positional displacements while neglecting probabilistic modeling of movement patterns. This approach fails to capture the intrinsic dynamics and semantic context of moving objects. In contrast, instead of just capturing surface-level displacements, the “motion intention” of human signals implicitly models visual features and latent motion distribution as shown in Fig.~\ref{framework_comparison} (b), thereby reducing the degradation in reconstruction quality caused by motion estimation errors. Fig.~\ref{framework_comparison} (c) provides visual example comparisons of the explicit or implicit motion paradigms, demonstrating that the explicit motion estimation mechanism can easily cause motion estimation distortions in human videos with complex deformation and large-scale motion.

To address the shortcomings of explicit motion-based approaches in Generative Human Video Compression (GHVC), this paper investigates the potential of Implicit Motion Transformation (IMT). In particular, we propose a novel framework that characterizes complex human body signals into compact features, subsequently transforming these features into implicit motion guidance to reconstruct human body signal. By shifting the focus from explicit to implicit motion transformation, the proposed approach could significantly improve reconstruction quality while minimizing artifacts for complex human video compression. Through a comprehensive study, we demonstrate the effectiveness of IMT in enhancing GHVC performance, ultimately contributing to advancements in video compression technology for dynamic human activities.

\section{The Proposed IMT Paradigm}

As illustrated in Fig.~\ref{fig2}, our proposed IMT framework enables efficient low-bandwidth communication for human body video. Specifically, at the encoder side, a key-reference frame capturing the human body's texture information is compressed using the state-of-the-art Versatile Video Coding (VVC) standard to serve as a texture reference for downstream signal synthesis. The subsequent inter frames are then processed through a neural network-based feature extractor, such that the high-dimensional human signal can be represented with compact visual features at a size of 6$\times$6$\times$1. These highly compact features undergo inter-prediction to reduce feature redundancy, followed by quantization and entropy coding to produce the final transmitted bitstream. 

Upon receiving the bitstream, the proposed decoder will perform the implicit motion transformation and vivid signal synthesis. The process begins with decoding the key-reference frame using the VVC codec, followed by two critical operations: appearance feature deformation for texture reference and compact feature extraction for feature compensation. Besides, the compact feature of inter frames can be obtained by entropy decoding and feature compensation. Subsequently, the decoded compact features from both key-reference and inter-frames are fed into feature up-sampling blocks, which can obtain motion-aware features with larger spatial size. Afterwards, the appearance feature and these motion features will be jointly input into a cross-attention transformer module to obtain transformed features that can describe the complex dynamic motion and clear texture/color information. Finally, relying on the strong inference capabilities of deep generative models, the human body video can be high-quality reconstructed from these transformed features. 

\subsection{Compact Visual Feature Extraction \& Encoding}

Analogous to CFTE~\cite{CHEN2022DCC}, the high-dimensional signal (\textit{i.e.,} the VVC reconstructed key-reference frame $\hat{K}$ or the subsequent inter frames $I_{l} \left (1\le l \le n , l\in Z   \right ) $) could be characterized into compact feature representation of temporal dynamics in a self-supervised learning manner.
This process involves three sequential stages: (1) spatial downsampling, (2) end-to-end feature learning, and (3) nonlinear normalization. 
Initially, $\hat{K}$ and $I_{l}$ are downsampled using a scale factor $s$ to reduce spatial redundancy. 
Subsequently, the downsampled frames are processed through an end-to-end U-Net network~\cite{RFB15a}, which learns hierarchical feature representations by combining encoder-decoder pathways with skip connections.
Finally, the extracted features are highly compacted into a latent code with the size of 6$\times$6$\times$1 via convolutional operations and enhanced with Generalized Divisive Normalization (GDN)~\cite{J2015Density}, where such a nonlinear transformation could improve feature expressiveness while suppressing irrelevant noise. 
The resulting compact feature representation $ \theta_{comp}$ can be defined as,
\begin{equation}
 \theta_{comp}= \varrho_{\left (conv,GDN  \right )}\left ( f_{U-Net}\left ( \phi \left ( X,s \right ) \right ) \right ),
\end{equation}  where $\phi\left ( \cdot \right )$, $f_{U-Net}\left ( \cdot \right )$ and $\varrho_{\left (conv,GDN  \right )} \left ( \cdot \right )$ represent down-sample operation, U-Net feature learning and nonlinear transformation, respectively. Herein, $X$ could be $\hat{K}$ or $I_{l}$.

To further enhance compression efficiency for compact features at a size of 6$\times$6$\times$1, a temporal prediction mechanism is implemented by performing inter-frame prediction between the current frame's feature and the previously reconstructed frame's feature. These resulting feature residuals across adjacent frames are then quantized and compressed using context-adaptive arithmetic coding~\cite{TEUHOLA1978308,1096090}, which can dynamically adjust probability models based on context priors to maximize bitstream efficiency. 

\subsection{Implicit Motion Transformation with Cross-Attention Transformer}

To enhance awareness of temporal dynamics in video reconstruction, we propose an implicit motion transformation module based on a cross-attention transformer architecture. Inspired by~\cite{10656311}, our design leverages cross-attention transformer mechanisms to explore the implicit evolution from compact feature representations to pixel-wise transformed features, thereby modeling temporal dependencies and enhancing the understanding of object motion dynamics.   

Specifically, at the decoder side, the key-reference frame $\hat{K}$ is decoded using the VVC codec, followed by two critical operations: (1) appearance feature deformation, where the decoded frame is processed through feature learning blocks to derive appearance features $f_{app}$, and (2) compact feature extraction, where a compact feature $\theta^{\hat{K}}_{comp}$ is extracted through a dedicated extractor. Besides, the compact feature of inter frames $\hat{\theta}^{I_{l}}_{comp}$ is decoded through arithmetic decoding and feature compensation operations. Subsequently, $\theta^{\hat{K}}_{comp}$ and $\hat{\theta}^{I_{l}}_{comp}$ are passed through feature up-sampling modules, enabling the generation of motion-aware features (\textit{i.e.,} $f^{\hat{K}}_{motion}$ and $f^{I_{l}}_{motion}$) with expanded spatial dimensions. 

To better ensure that subtle temporal variations are preserved during the feature motion transformation process, we introduce the cross-attention transformer to dynamically aligns motion patterns across feature dimensions. Herein, $f_{app}$, $f^{\hat{K}}_{motion}$ and $f^{I_{l}}_{motion}$ will be further projected into the corresponding value feature $f_{V}$, key feature $f_{K}$ and query feature $f_{Q}$ to perform the cross attention operation. The attention-modulated feature  $f^{I_{l}}_{atten}$ can be formulated by,
\begin{equation}
f^{I_{l}}_{atten}=\mathit{Softmax} \left ( f_{Q},\left (  f_{K}\right )^{T}   \right ) \times f_{V} + f^{I_{l}}_{motion},
\end{equation} where $\mathit{Softmax}\left ( \cdot \right )$ is a softmax normalization function.
Afterwards, $f^{I_{l}}_{atten}$ and $f^{I_{l}}_{motion}$ are input to a Transformer block $\tau $, where they undergo normalization and feed-forward operations. This process yields a transformed feature $f^{I_{l}}_{trans}$ that encodes enhanced contextual relationships, effectively integrating temporal and spatial information for downstream signal reconstruction task. The specific process can be represented as, 
\begin{equation}
f^{I_{l}}_{trans}=\tau \left ( f^{I_{l}}_{atten}, f^{I_{l}}_{motion} \right ).
\end{equation}

\begin{figure}[t]
\centering
\vspace{-0.3em}
\centerline{\includegraphics[width=0.5 \textwidth]{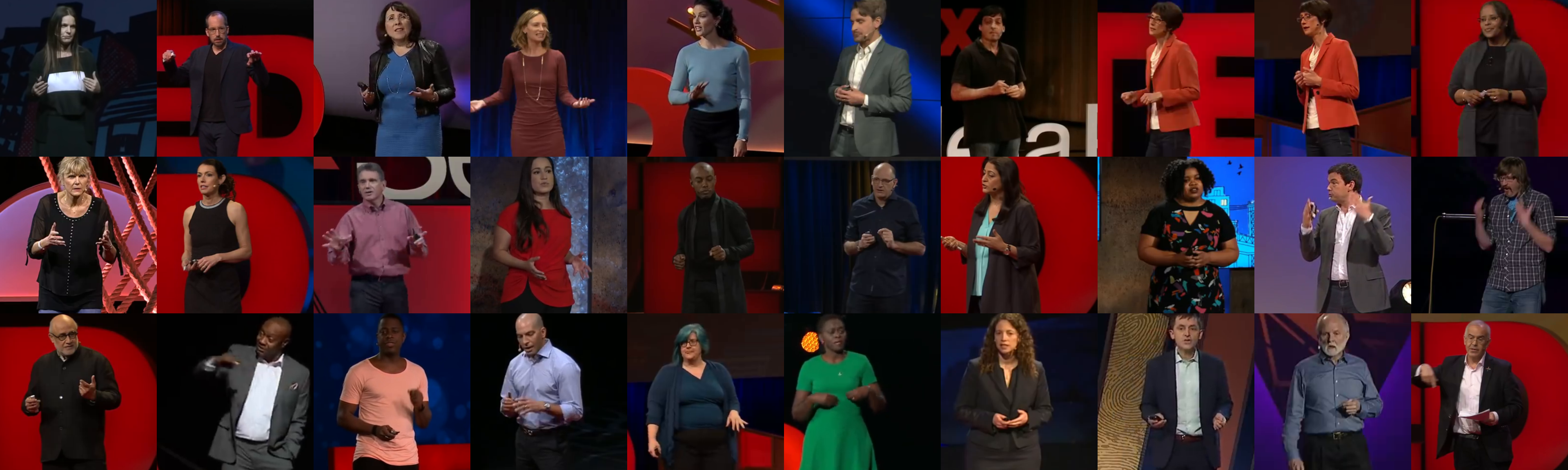}}
\caption{Illustration of 30 testing sequences selected and pre-processed from TED-Talk dataset~\cite{siarohin2021motion}.} 
\vspace{-1.5em}
\label{fig3}
\end{figure}

\subsection{GAN-based Human Frame Generation}

Deep generative networks leverage their robust inference capabilities to enable realistic signal reconstruction within the generative compression framework. Herein, the GAN architecture~\cite{goodfellow2014generative} is adopted due to its demonstrated advantages in computational efficiency and deployment scalability compared to other generative models like Diffusion Models~\cite{NEURIPS2021_49ad23d1}.
In contrast to conventional flow-based frame generation approaches~\cite{siarohin2021motion,CHEN2022DCC,yin2024generative,FV2V} that rely on dense motion flow to warp the key-reference frame for the generation of inter frames, this work introduces a direct generation strategy to synthesize realistic human signals from the transformed features $f^{I_{l}}_{trans}$, bypassing explicit motion modeling and guidance. This approach mitigates motion error propagation while retaining clear texture details, thereby enhancing visual fidelity in generated outputs. The overall process can be described as,
\begin{equation}
{
{\hat{I}}_{l} =  G_{frame} \left (f^{I_{l}}_{trans}  \right )
},
\end{equation} where $G_{frame}\left ( \cdot \right )$ represents the generator's network layers that convert transformed features into human body frames. A multi-scale feature discriminator~\cite{wang2018pix2pixHD} is  employed to ensure that each generated frame ${\hat{I}}_{l}$ achieves realistic reconstruction under the supervision of ground-truth images $I_{l}$.

\subsection{Model Optimization}
Herein, we employ an end-to-end training strategy to jointly optimize the compact feature extraction, implicit motion transformation and GAN-based human frame generation modules. The training objective is formulated by combining three complementary losses: (1) a perceptual similarity loss $\mathcal L_{per}$~\cite{johnson2016perceptual} to preserve high-level semantic consistency, (2) a discriminator-guided adversarial loss $\mathcal L_{adv}$~\cite{wang2018pix2pixHD} for photo-realistic synthesis, and (3) a texture fidelity loss $\mathcal L_{tex}$~\cite{7780634} to ensure fine-grained appearance preservation across generated frames. The detailed formulation is listed as follows,
\begin{equation}
\mathcal L_{total}=\lambda _{per}\mathcal L _{per}+\lambda _{adv}\mathcal L _{adv}+\lambda _{tex}\mathcal L _{tex},
\end{equation} where $\lambda_{per}$, $\lambda _{adv}$ and $\lambda _{tex}$ are set to 10, 1 and 1000, respectively.

\begin{figure*}[!t]
\centering
\vspace{-4em}
\subfloat[Rate-DISTS]{\includegraphics[width=0.333 \textwidth]{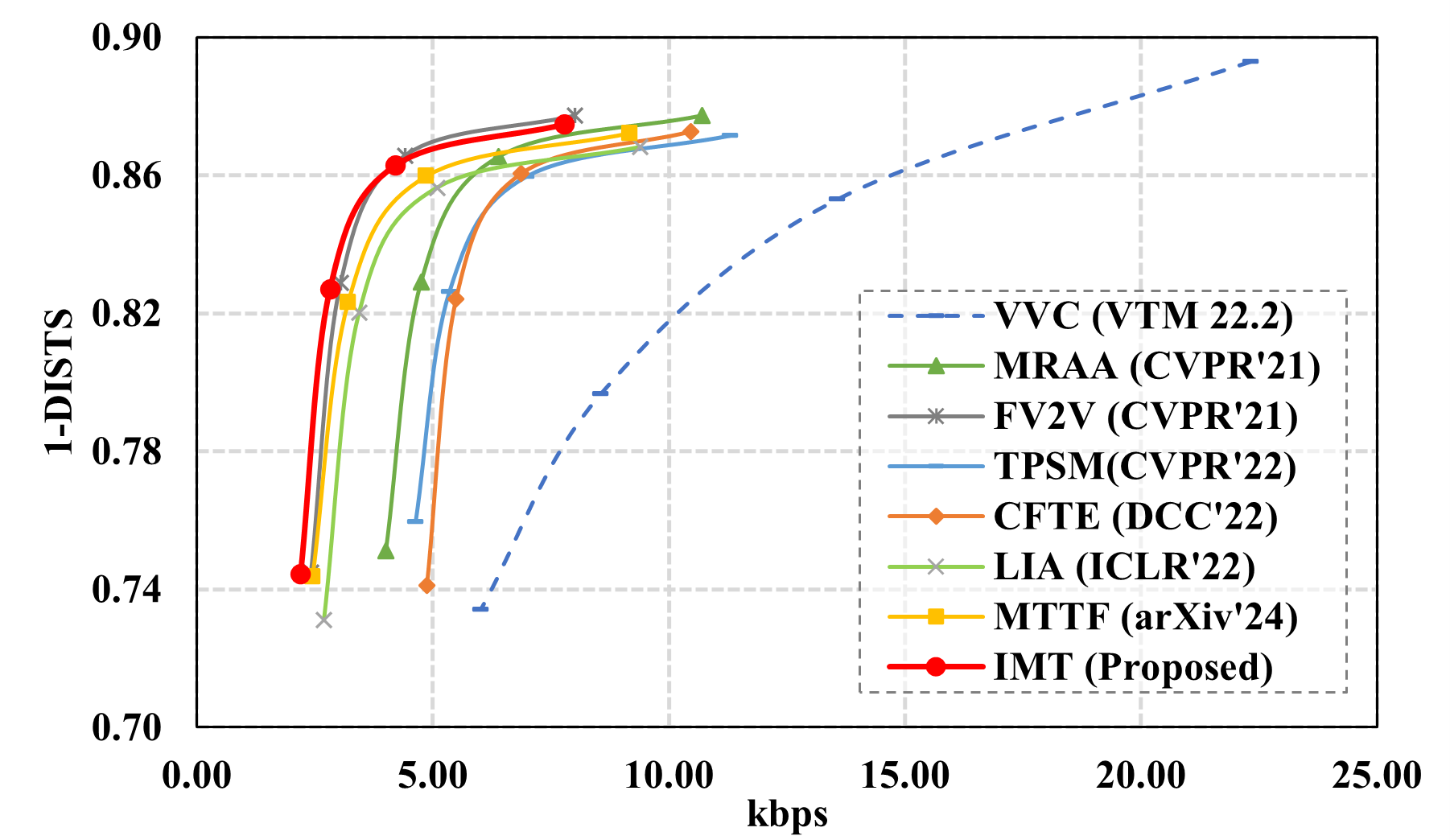}}
\subfloat[Rate-LPIPS]{\includegraphics[width=0.333 \textwidth]{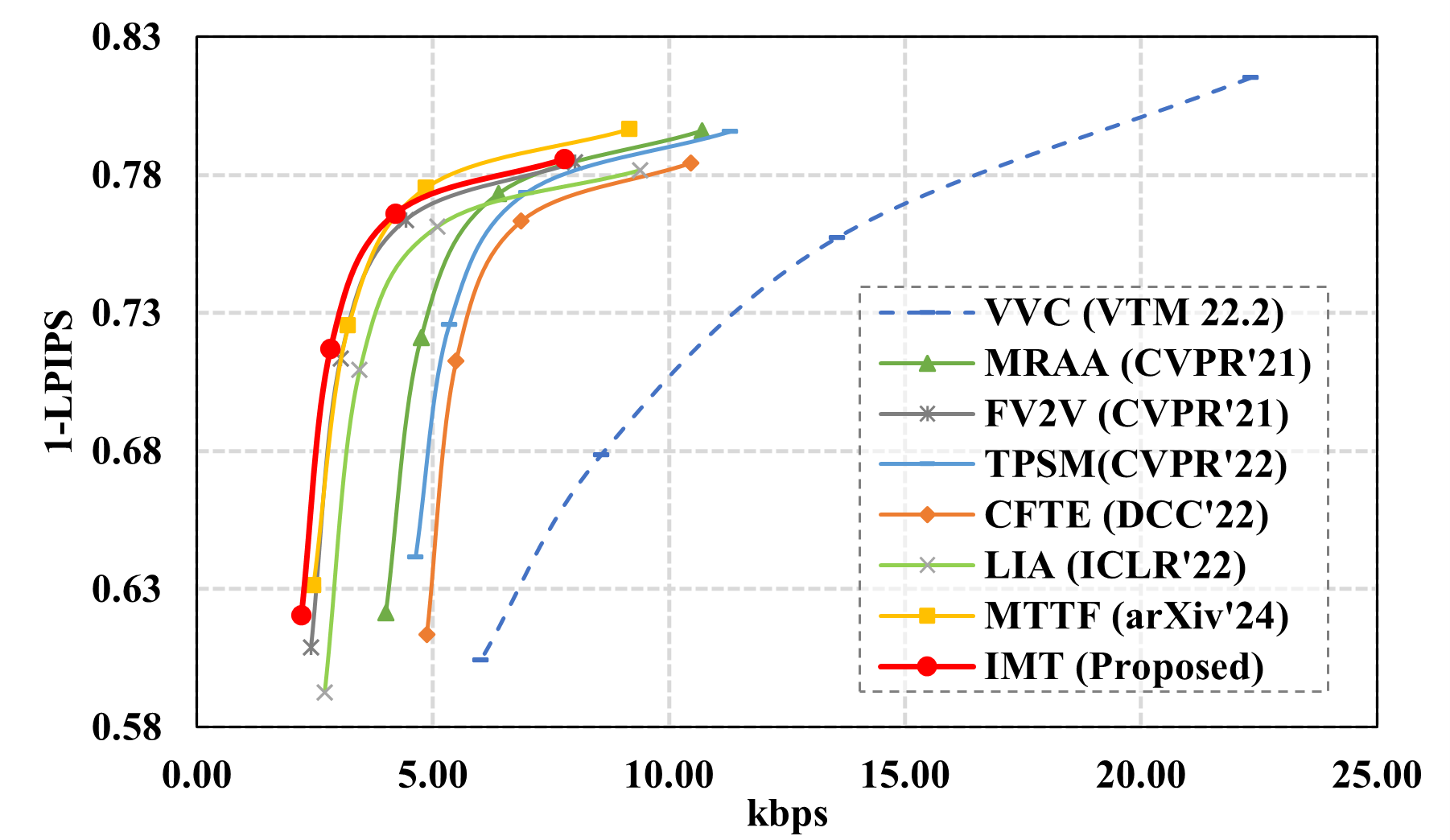}}
\subfloat[Rate-FVD]{\includegraphics[width=0.333 \textwidth]{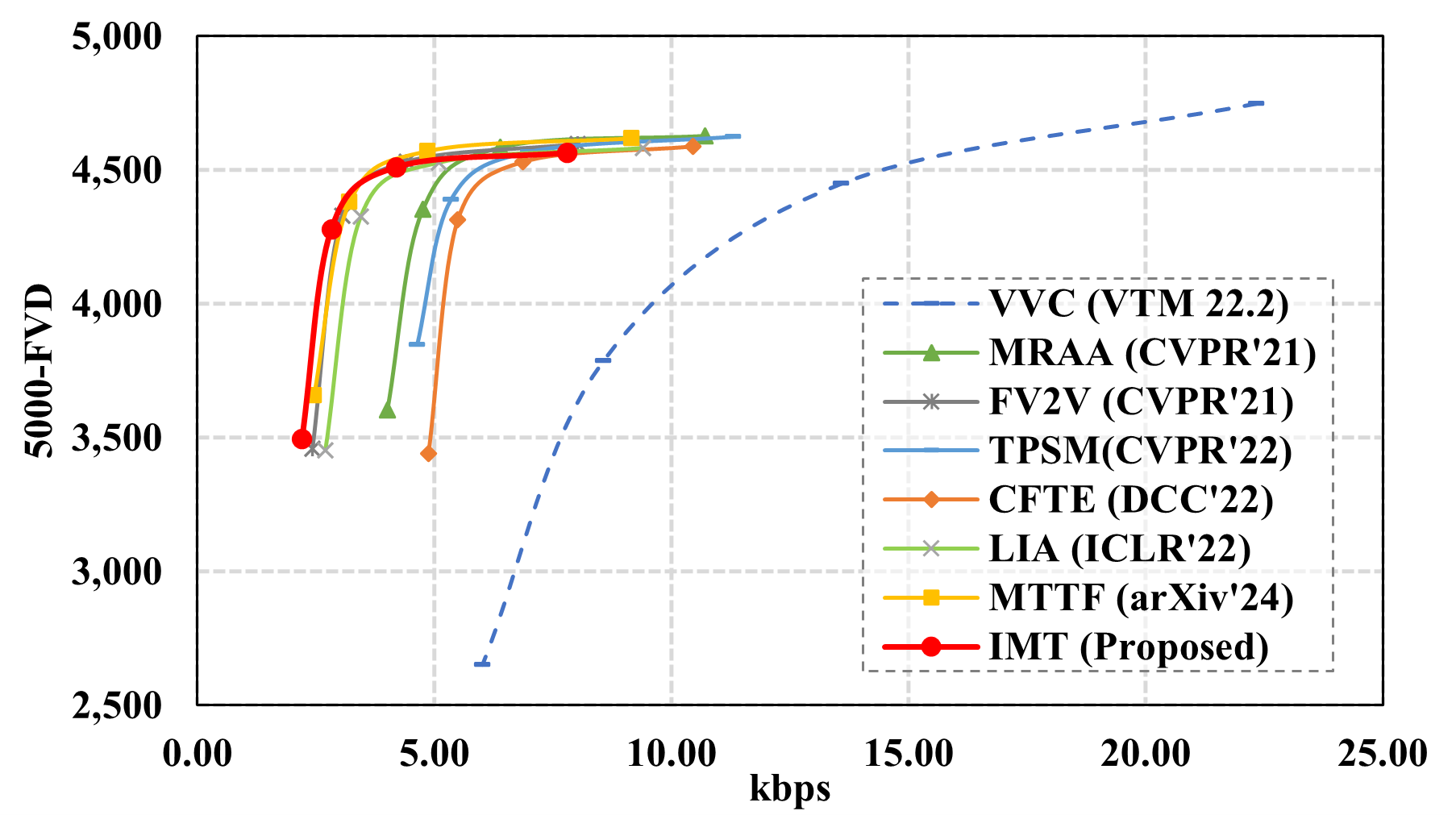}}
\caption{Overall RD performance comparisons with VVC~\cite{bross2021overview}, MRAA~\cite{siarohin2021motion}, FV2V~\cite{FV2V}, TPSM~\cite{zhao2022thin}, CFTE~\cite{CHEN2022DCC}, LIA~\cite{wanglatent} and MTTF~\cite{yin2024generative} in terms of rate-DISTS, rate-LPIPS and rate-FVD. }
\label{fig_RD}  
\vspace{-0.8em}
\end{figure*}

\begin{figure*}[!t]
\centering
\subfloat[Similar Bitrate Comparisons]{\includegraphics[width=1 \textwidth]{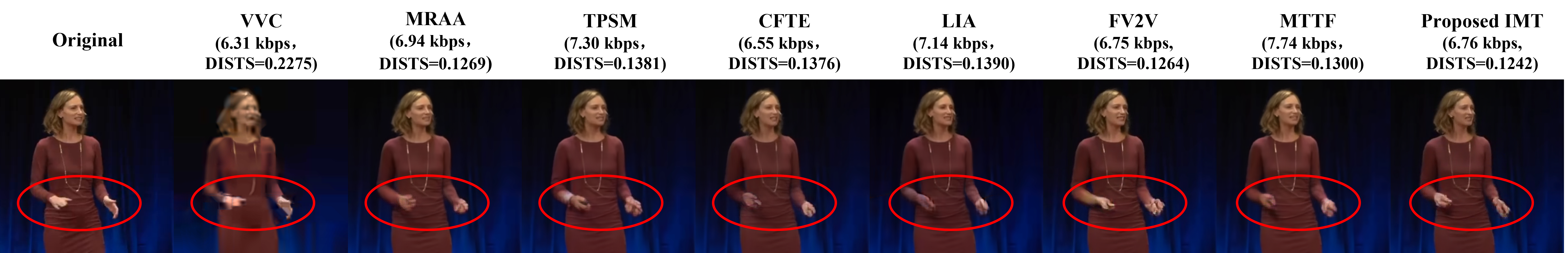}} \\
\vspace{-0.8em}
\subfloat[Similar Quality Comparisons]{\includegraphics[width=1 \textwidth]{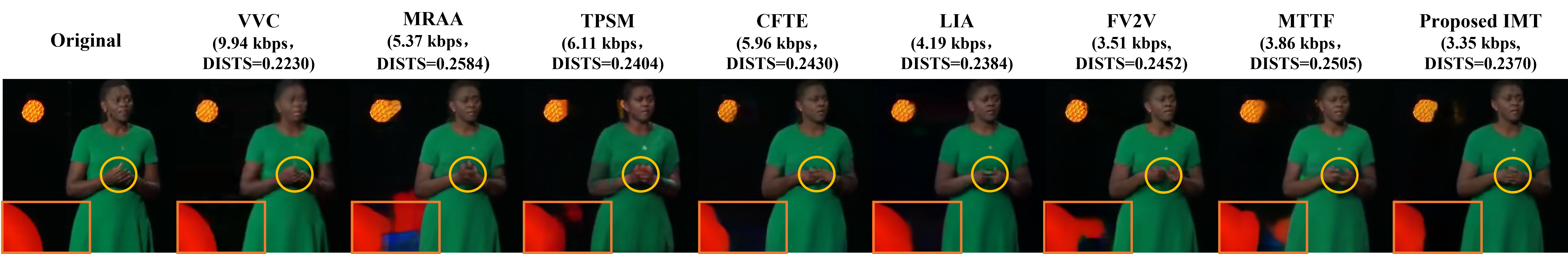}}
\caption{Visual quality comparisons among VVC~\cite{bross2021overview}, MRAA~\cite{siarohin2021motion}, FV2V~\cite{FV2V}, TPSM~\cite{zhao2022thin}, CFTE~\cite{CHEN2022DCC}, LIA~\cite{wanglatent}, MTTF~\cite{yin2024generative} and IMT (Proposed) at the similar bitrate and quality. Lower DISTS scores indicate higher perceptual quality. }
\label{fig_sub}  
\vspace{-1.5em}
\end{figure*}

\section{Experimental Results}

\subsection{Experimental Settings}
\textbf{Implementation Details.}
The proposed model is implemented using the PyTorch libraries and trained on the TED-Talk dataset~\cite{siarohin2021motion}, which consists of 1,132 training videos and 128 testing videos at a resolution of 384$\times$384. Training is conducted over 300 epochs with 30 times data repeat strategy on 4 NVIDIA TESLA V100 GPUs. For optimization, the Adam optimizer is configured with $\beta _{1}$ = 0.5 and $\beta _{2}$= 0.999, while a learning rate of 0.0002 is adopted to balance convergence stability and training speed.

\textbf{Compared Algorithms.}
To demonstrate the effectiveness of the proposed IMT paradigm, we compare the state-of-the-art video coding standard VVC~\cite{bross2021overview} and six representative generative human video compression algorithms, including MRAA~\cite{siarohin2021motion}, FV2V~\cite{FV2V}, TPSM~\cite{zhao2022thin}, CFTE~\cite{CHEN2022DCC}, LIA~\cite{wanglatent} and MTTF~\cite{yin2024generative} that rely on explicit motion estimation strategies. 
These testing algorithms are re-trained with the same training settings in the TED-Talk training dataset~\cite{siarohin2021motion} for fair comparisons. We select 30 testing sequences from TED-Talk testing dataset, where each sequence contains 150 frames with a resolution of 384$\times$384. 
To better compare the performance, we follow the JVET GFVC AhG test conditions~\cite{JVET-AJ2035}.
The specific settings of these compared algorithm are provided as follows,
\begin{itemize}
\item{\textbf{VVC Anchor:} The VTM-22.2 reference software, corresponding to the VVC standard, is employed as the anchor. The encoder is configured in the Low-Delay-Bidirectional (LDB) mode with the Quantization Parameters (QPs) of \{37, 42, 47, 52\}. All raw video inputs are converted to the standardized YUV 4:2:0 chroma format.}
\item{\textbf{Generative Codecs:} The key-reference frame of the raw video is first converted to the YUV 4:2:0 chroma format and then compressed using the VTM-22.2 reference software with the QPs of \{22, 32, 42, 52\}. The subsequent inter frames will be represented into compact parameters, which can be further compressed via a context-adaptive arithmetic coder.}
\end{itemize}

\textbf{Evaluation Measures.} 
As reported in~\cite{10477607,GFVC_Survey}, compared with traditional pixel-wise quality metrics such as PSNR and SSIM, learning-based perceptual quality measures like DISTS~\cite{DISTS}, LPIPS~\cite{LPIPS} and FVD~\cite{unterthiner2019fvd} are more suitable for evaluating generative video compression due to their strong ability to capture perceptual quality in learned representation spaces. 
Therefore, we leverage these three perceptual metrics as evaluation criteria to assess the perceptual quality of reconstructed human body videos. Lower DISTS/LPIPS/FVD scores correlate with better perceptual quality. As such, Fig. \ref{fig_RD} uses 1-DISTS, 1-LPIPS and 5000-FVD to present more coherent RD curves. Furthermore, we incorporate the encoding bitrate (measured in kilobits per second, or kbps) as a quantitative metric, such that the Rate-Distortion curve (\textit{i.e.,} RD-curve) and {Bjøntegaard}-Delta-rate (\textit{i.e.,} BD-rate)~\cite{Bjntegaard2001CalculationOA} can be summarized to assess compression efficiency.

\begin{table}[t]
\vspace{-1.3em}
\renewcommand\arraystretch{1.2}
\caption{Average Bit-rate savings of 30 testing sequences in terms of Rate-DISTS, Rate-LPIPS and Rate-FVD }  
\label{table1}
\centering
\resizebox{0.46\textwidth}{!}{
\begin{tabular}{cccc}
\hline
Anchor: VVC (VTM 22.2)  & Rate-DISTS        & Rate-LPIPS        & Rate-FVD          \\ \hline
MRAA (CVPR'21)          & -50.92\%          & -51.02\%          & -55.66\%          \\
FV2V (CVPR'21)          & -68.49\%          & -67.18\%          & -70.79\%          \\
TPSM(CVPR'22)           & -44.51\%          & -47.22\%          & -52.08\%          \\
CFTE (DCC'22)           & -61.99\%          & -61.77\%          & -66.94\%          \\
LIA (ICLR'22)           & -40.33\%          & -40.51\%          & -44.73\%          \\
MTTF (arXiv'24)         & -65.96\%          & -68.16\%          & -71.27\%          \\
\textbf{IMT (Proposed)} & \textbf{-70.52\%} & \textbf{-70.56\%} & \textbf{-72.46\%} \\ \hline
\end{tabular}
}
\vspace{-1.3em}
\end{table}

\subsection{Performance Comparisons}

\textbf{Rate-Distortion Performance.} Fig. \ref{fig_RD} presents a comparative analysis of the RD performance of our proposed IMT compression framework against VVC and six generative compression approaches, evaluated across three perceptual quality metrics on the TED-Talk test dataset. The results demonstrate that our method achieves substantial bitrate savings relative to the state-of-the-art VVC standard, particularly in learning-based quality assessment metrics. When benchmarked against competitive algorithms including MRAA, TPSM, CFTE, and LIA, our IMT framework maintains a consistent performance advantage. Notably, compared to CFTE—which employs explicit motion estimation with equivalent feature dimensions, our implicit motion transformation strategy demonstrates significant RD performance improvements, highlighting the efficacy of the proposed motion modeling approach. In comparison with FV2V and MTTF algorithms, the IMT scheme exhibits superior performance across most bitrate ranges, though a marginal performance degradation is observed at high bitrates. 

Table \ref{table1} quantifies the bit-rate savings of our proposed IMT scheme across 30 test sequences, demonstrating significant reductions compared to the VVC anchor. In particular, the proposed IMT can achieve 70.52\% bit-rate savings in terms of DISTS, 70.56\% bit-rate savings in terms of LPIPS and 72.46\% bit-rate savings in terms of FVD  compared with the VVC anchor, outperforming all existing generative compression algorithms.

\textbf{Subjective Quality.} Fig. \ref{fig_sub} provides visual quality comparisons at similar bitrates and similar quality among different compression schemes. The proposed IMT demonstrates superior reconstruction capabilities compared to the VVC codec, achieving enhanced visual fidelity with significantly reduced artifacts in human video sequences. Furthermore, when benchmarked against other generative compression frameworks, the proposed IMT excels in preserving fine local details while effectively suppressing background interference, facilitating perceptually more accurate and coherent reconstructions. 
 

\section{Conclusions}
This paper addresses the critical challenge of Generative Human Video Coding (GHVC) by identifying the limitations of explicit motion-based approaches in handling complex human body dynamics. By proposing the Implicit Motion Transformation (IMT) paradigm, we demonstrate the effectiveness of shifting from explicit motion supervision to implicit motion modeling. The key innovation lies in encoding high-dimensional human body signals into compact visual features and leveraging these features to infer implicit motion patterns for signal reconstruction. Our work emphasizes the importance of rethinking motion representation and transformation paradigms for complex dynamic scenes, and develops a new cross-attention transformer architecture for implicit motion learning.
This approach not only mitigates severe distortions caused by explicit motion estimation errors but also achieves state-of-the-art compression efficiency and fidelity in human body video coding, offering a robust framework for future research in generative video compression.

\bibliographystyle{IEEEtran}
\bibliography{main}
\end{document}